\documentclass[lettersize,journal]{IEEEtran}

\usepackage{lineno,hyperref}
\usepackage{colortbl}
\usepackage{amsmath,amssymb,amsfonts}
\usepackage{algorithmic}
\usepackage{graphicx}
\usepackage{float}
\usepackage{makecell}
\usepackage{amsmath,amsfonts}
\usepackage{algorithmic}
\usepackage{algorithm}
\usepackage{array}
\usepackage[caption=false,font=normalsize,labelfont=sf,textfont=sf]{subfig}
\usepackage{textcomp}
\usepackage{stfloats}
\usepackage{url}
\usepackage{verbatim}
\usepackage{graphicx}
\usepackage{cite}
\AtBeginDocument{%
  }
\usepackage{multirow}
\usepackage{graphicx}
\usepackage{amsmath}
\usepackage{booktabs}
\usepackage{color}
\usepackage{picins}

\begin{document}

\title{Unstoppable Attack: Label-Only Model Inversion via Conditional Diffusion Model}

\author{Rongke Liu, Dong Wang, Yizhi Ren, Zhen Wang, Kaitian Guo, Qianqian Qin, Xiaolei Liu
        % <-this % stops a space~\IEEEmembership{Staff,~IEEE,
\thanks{Manuscript received 1 August 2023; revised 4 December 2023 and 14 February 2024; accepted 24 February 2024. The associate editor coordinating the review of this manuscript and approving it for publication was Prof. Tondi, Benedetta. \textit{(Corresponding author: Dong Wang.)}}%
\thanks{Rongke Liu, Dong Wang, Yizhi Ren, Zhen Wang, Kaitian Guo and Qianqian Qin are with the School of Cyberspace, Hangzhou Dianzi University, Hangzhou 310018, China (e-mail: liurk@hdu.edu.cn; wangdong@hdu.edu.cn; renyz@hdu.edu.cn; wangzhen@hdu.edu.cn; guokt@hdu.edu.cn; qinqq@hdu.edu.cn)}%
\thanks{Xiaolei Liu are with the Institute of Computer Application, China Academy of Engineering Physics, Mianyang 621900, China (e-mail: luxaole@gmail.com)}%
\thanks{Digital Object Identifier 10.1109/TIFS.2024.3372815}
}

\markboth{IEEE TRANSACTIONS ON INFORMATION FORENSICS AND SECURITY,~Vol.~x, No.~x, x~2024}%
{Shell \MakeLowercase{\textit{et al.}}: A Sample Article Using IEEEtran.cls for IEEE Journals}


\maketitle

\begin{abstract}
Model inversion attacks (MIAs) aim to recover private data from inaccessible training sets of deep learning models, posing a privacy threat. MIAs primarily focus on the white-box scenario where attackers have full access to the model’s structure and parameters. However, practical applications are usually in black-box scenarios or label-only scenarios, i.e., the attackers can only obtain the output confidence vectors or labels by accessing the model. Therefore, the attack models in existing MIAs are difficult to effectively train with the knowledge of the target model, resulting in sub-optimal attacks. To the best of our knowledge, we pioneer the research of a powerful and practical attack model in the label-only scenario.

In this paper, we develop a novel MIA method, leveraging a conditional diffusion model (CDM) to recover representative samples under the target label from the training set. Two techniques are introduced: selecting an auxiliary dataset relevant to the target model task and using predicted labels as conditions to guide training CDM; and inputting target label, pre-defined guidance strength, and random noise into the trained attack model to generate and correct multiple results for final selection. This method is evaluated using Learned Perceptual Image Patch Similarity as a new metric and as a judgment basis for deciding the values of hyper-parameters. Experimental results show that this method can generate similar and accurate samples to the target label, outperforming generators of previous approaches.
\end{abstract}


\maketitle

\section{Introduction}
\IEEEPARstart{T}he technology of artificial intelligence is developing rapidly and its application brings many conveniences to our daily life nowadays. For example, in the field of image recognition, deep neural networks (DNNs) assist in applications such as face and fingerprint recognition, biomedical diagnosis, and extracting useful information from massive datasets of images. However, building such a model requires massive data for training, which may contain private or sensitive information. Some studies have noted that models tend to memorize the training data\cite{song2017machine}\cite{tramer2016stealing}\cite{papernot2018sok}\cite{he2020towards}.

The model inversion attack (MIA) is a type of privacy attack that aims to regenerate data to represent training data, input data, or sensitive attributes by accessing the target model \cite{fredrikson2014privacy}\cite{fredrikson2015model}\cite{yang2019neural}. For instance, the face image of a target individual in the training set can be recovered by a face recognition model, similarly, the input face image can be reconstructed from the prediction vector produced by the face recognition model. Or sensitive attributes of an individual's genome can be inferred by a medical prediction model\cite{fredrikson2014privacy}. Unlike membership inference \cite{shokri2017membership} and model extraction attacks \cite{tramer2016stealing}, MIA concentrates on recovering data closely resembling the private data itself. Existing model inversion attacks can be classified into two types based on the attacker's background knowledge: the white-box and the black-box scenarios.

In the white-box scenario, the attacker has full access to the structure and parameters of the target model, while in the black-box scenario, the attacker can only access the predictions (confidence vectors or labels) of the target model without providing gradient or other information. For the face recognition model, the existing white-box attacks \cite{fredrikson2015model} iteratively optimize the input image by feeding a noise image to the target model and minimizing the loss between the prediction and target label, achieved through a gradient descent algorithm. Whereas, in the attack scenarios where the target model is usually a DNN (e.g., convolutional neural networks (CNNs)), the sensitive features to be recovered often lie in a high-dimensional, continuous data space. Directly optimizing over the high-dimensional space without any constraints may generate unrealistic features lacking semantic information\cite{fredrikson2015model}\cite{zhang2020secret}.   In order to obtain more semantic and meaningful images on CNNs\cite{lecun1998gradient}, state-of-the-art methods \cite{zhang2020secret}\cite{chen2021knowledge}\cite{yuan2023pseudo} generate the image by feeding noisy vectors to the generator in a generative adversarial network (GAN)\cite{goodfellow2020generative} trained with an auxiliary dataset, therefore the optimization turns to the noisy vectors and generator. 

In contrast, the existing black-box attack \cite{yang2019neural}\cite{zhu2022label} is to optimize a generator by gradient descent to minimize the pixel loss between the generated and auxiliary images. The generated images are produced by the aforementioned generator based on the predictions of the target model for the auxiliary data. In addition, based on the black-box model output confidence or label, the attacks can be categorized into “data reconstruction" and “training class inference" \cite{yang2019neural}. The “data reconstruction" is mainly based on the prediction confidence vector of the target model to recover the input samples \cite{yang2019neural}, while “training class inference" is based on the one-hot vectors or labels to recover the representative samples of the target \cite{yang2019neural}\cite{zhu2022label}. Since this paper studies label-only scenarios, our research belongs to the “training class inference".

\begin{figure}[h]
  \centering
  \includegraphics[width=0.7\linewidth]{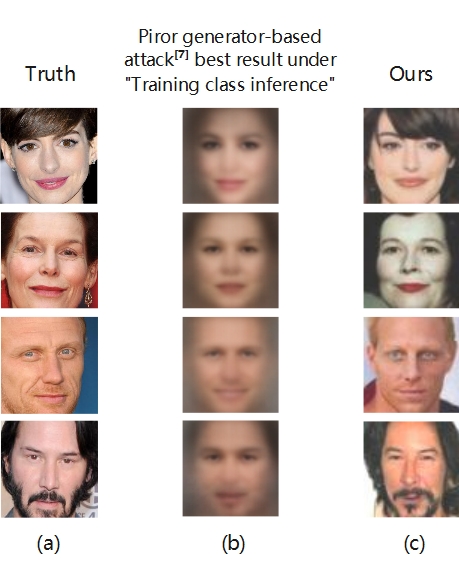}
  \caption{Training class inference of our and previous approaches against a facial recognition classifier in the label-only scenario. For better comparison with our method, we turned the labels into correct one-hot vectors to train (b)'s attack model\cite{yang2019neural} for recovering optimal color images. Note that the correct one-hot vector implies that the confidence value at the target position is 1, while the rest are 0. For instance, if there are a total of 3 classes and the target is class 1, then the one-hot vector would be (1,0,0). Moreover, the auxiliary dataset used by both is the same.}
  \label{F1}
\end{figure}

Current research yielding ideal generation results primarily focuses on white-box scenarios. However, models are typically accessed as black boxes in practical applications and only output predicted labels. This prevents the generator in the state-of-the-art white-box attack from optimizing with the help of gradient and prevents decoupling the potential space of the target class \cite{yuan2023pseudo}. Moreover, there are several limitations to existing black-box attacks: 
\begin{enumerate}
    \item The generated images \cite{dionysiou2023exploring}\cite{yang2019neural}\cite{zhu2022label}\cite{ye2022model}\cite{yoshimura2021model}\cite{fredrikson2015model} are mostly grey-scale, which cannot accurately determine the color characteristics of the target, such as skin tone or pupil color. 

    \item Since there is currently no attack model that can be optimally trained in the label-only scenario, existing methods \cite{kahla2022label}\cite{han2023reinforcement}\cite{an2022mirror} had to reach the attack goal by designing additional optimization strategies based on the generator from ordinary GAN, which is trained without the target model's knowledge.

    \item Only a single sample can be generated for the target label \cite{yang2019neural}.

    \item The generated results for the target label based on the generator are sub-optimal and evaluation metrics lack comprehensiveness. A comparison of the specific results and evaluation metrics can be seen in Figure \ref{F1} and Table \ref{Table 1}.
\end{enumerate}

In this paper, we develop a novel label-only model inversion attack method to address the above limitations. The core design idea of our method is to train a conditional diffusion model (CDM)\cite{ho2022classifier} guided by the target model predicted label, and in the recovery phase, various samples can be recovered for selection based on the target label guidance. Since our attack method only needs the target model to predict the labels and existing defensive strategies\cite{wang2021improving}\cite{peng2022bilateral}\cite{yang2023purifier}\cite{struppek2023careful} should ensure the availability of the model, this attack is unstoppable. Table \ref{Table 2} visualizes the strengths and uniqueness of our study.

\begin{table}[t]
\centering
\label{Table 1}
\caption{Comparison of evaluation metrics in work related to black-box model inversion attacks, where $\bigstar$ means that the work includes this metric.}
\setlength{\tabcolsep}{4mm}{
\begin{tabular}{@{}cccl@{}}
\toprule
              & \begin{tabular}[c]{@{}c@{}}Attack \\ accuracy\end{tabular} & \begin{tabular}[c]{@{}c@{}}Feature/Pixel-level \\ similarity\end{tabular} & \begin{tabular}[c]{@{}l@{}}Perceptual  similarity/\\Quantification of\\ qualitative assessment
 \end{tabular}\\ \midrule

{\cite{yoshimura2021model}}      &                                                           & $\bigstar$                                                   &                                                                 \\ \midrule
{\cite{zhu2022label}}      &                                                           & $\bigstar$                                                   &                                                                 \\
\midrule
{\cite{kahla2022label}}      & $\bigstar$                                   &                                                                           &                                                                 \\ \midrule
{\cite{dionysiou2023exploring}}       & $\bigstar$                                  & $\bigstar$                                                   &                                                                 \\ \midrule
{\cite{han2023reinforcement}}       & $\bigstar$                                   & $\bigstar$                                                   &                                                                 \\ \midrule
\textbf{Ours} & $\bigstar$                                   & $\bigstar$                                                   & \multicolumn{1}{c}{$\bigstar$}                     \\ \bottomrule
\end{tabular}}%
\end{table}

Specifically, we first select an auxiliary dataset that is relevant to the target model task. For example, if the training set utilized by the target model is a facial dataset, then the auxiliary dataset will correspondingly be facial. Secondly, we assign the predicted labels produced by the target model for the corresponding auxiliary data. These predicted labels can reflect the target model's judgments about various types of target features in the training set. Then, we train the CDM, which consists of forward diffusion and backward prediction \cite{ho2020denoising}, to add Gaussian noise to the auxiliary data in the forward diffusion process. This process eventually makes the image close to the standard normal distribution noise image. In the backward prediction process, the added noise is predicted under the guidance of the predicted label, which allows the diffusion model to learn the noise distribution added by the target under the prediction label. After training, we feed random standard normal distribution noise images and the target label into the CDM to recover images with a pre-defined guidance strength. Due to the difference with the traditional CDM training process, the auxiliary data of this model for a specific target comes from public data, and the real labels do not correspond to the actual data, which makes the difference in the learned noise distribution and leads to poor noise reduction in the generated images. Accordingly, we perform gamma correction \cite{hunt2005reproduction} on the generated images to make them more consistent with human visual judgment. Finally, we randomly change multiple generated images, submit them to the target model for prediction, and then select the top-$k$ robust generated images. Experiments show that our attack model can generate more accurate, realistic, and similar images than existing attack models in the label-only scenario.

\begin{table*}[]
\centering
\caption{Strengths and uniqueness of our study compared to related work, where $\bigstar$ means that the work includes this Capability.}
\label{Table 2}
\setlength{\tabcolsep}{2.4mm}{%
\begin{tabular}{@{}ccccccc@{}}
\toprule
\multirow{2}{*}{Capability}                                                                & \multicolumn{2}{c}{White-box} & \multicolumn{4}{c}{Black-box}                        \\ \cmidrule(l){2-7} 
 & \begin{tabular}[c]{@{}c@{}}GMI\\\cite{zhang2020secret}\end{tabular}& \begin{tabular}[c]{@{}c@{}}PLG\\\cite{yuan2023pseudo}\end{tabular}           &\begin{tabular}[c]{@{}c@{}}BERP\\\cite{kahla2022label}\end{tabular}     & \begin{tabular}[c]{@{}c@{}}RL-MIA\\\cite{han2023reinforcement}\end{tabular}      & \begin{tabular}[c]{@{}c@{}}LB-MIA\\\cite{yang2019neural}\end{tabular}      & \begin{tabular}[c]{@{}c@{}}\textbf{Ours}\end{tabular} \\ \midrule
Access to the target model only output label                                               &               &               & $\bigstar$ &            &            & $\bigstar$    \\ \midrule
Attack methods focus on generative models                                                  & $\bigstar$    & $\bigstar$    &            &            & $\bigstar$ & $\bigstar$    \\ \midrule
No need to obtain gradient information from the target model to attack &               &               & $\bigstar$ & $\bigstar$ & $\bigstar$ & $\bigstar$    \\ \midrule
The generator does not require additional optimization strategies to achieve targeted attacks                    &      & $\bigstar$    &            &            & $\bigstar$ & $\bigstar$    \\ \bottomrule
\end{tabular}%
}
\end{table*}

As shown in Table \ref{Table 1}, in the field of MIA, evaluation metrics are not standardized, and commonly used measures such as attack accuracy and feature distance may not accurately reflect the quality of generated results. For instance, an over-fitted and semantically meaningless optimized image may still exhibit high attack accuracy and low feature distance. We argue that a qualitative evaluation may be more important than a quantitative one for this work \cite{fredrikson2015model}\cite{struppek2022plug}\cite{dionysiou2023exploring}. As such, we propose the use of Learned Perceptual Image Patch Similarity (LPIPS) \cite{zhang2018unreasonable} as a novel evaluation metric for MIA. LPIPS approximates human judgment of similarity between two sets of data and can serve as a proxy for qualitative evaluation. Our results demonstrate that our method is capable of generating accurate and similar data to target labels in the label-only scenario and outperforms the previous attack models regardless of the individual overlap between the auxiliary set and the training set.

{\bfseries Contributions. }In summary, we make the following contributions to this paper:
\begin{itemize}
\item We pioneer a new MIA attack model that effectively leverages target model knowledge even in label-only scenarios. Our experiments validate the practicality and effectiveness of our approach.

\item We propose to utilize gamma correction to address the degradation of generation quality due to differences in auxiliary data under the same target.

\item We can use the target model to filter out multiple generated samples under the same target. Compared with the existing methods in the label-only scenario, which can only generate a unique sample for the same target, this method is more fault-tolerant and has a large optimization space.

\item We conduct a systematic evaluation of our attack in terms of attack accuracy, similarity, and realism, both quantitatively and qualitatively. Our experimental results demonstrate that our attack model can generate more accurate, realistic, and similar target samples in the label-only scenario, regardless of the individual overlap between the auxiliary set and the training set.
\end{itemize}

\section{Related Work and Preliminary Knowledge}
Privacy attacks on machine learning and deep learning models can be categorized into three types: model extraction attacks \cite{tramer2016stealing}, inference attacks \cite{shokri2017membership}, and model inversion attacks \cite{fredrikson2015model}. Model extraction attacks aim to infer the parameters or features of the target model to replicate a similar machine-learning model, while inference attacks aim to reveal information related to the training set of the target model. Model inversion attack aims to regenerate data to represent training data, input data, or sensitive attributes by accessing the trained target model. Among them, the MIA reveals privacy information at a finer level.

\subsection{Traditional Model Inversion Attacks}
Fredrikson et al. \cite{fredrikson2014privacy} were the first to propose a model inversion attack, using warfarin dose personalization\cite{international2009estimation} as a case study to show how MIA can infer patient-specific genetic markers by accessing linear regression models as well as maximum posterior probabilities. Subsequently, Fredrikson et al. \cite{fredrikson2015model} extended the attack to decision trees\cite{quinlan1986induction} and face recognition neural networks using confidence and gradient descent algorithms. However, both works explored black-box attacks, but \cite{fredrikson2014privacy} assumed too strongly on the knowledge held by the adversary, and \cite{fredrikson2015model} required 50-80 days for experiments on multi-layer perception network or denoising autoencoder network to complete, i.e., estimating gradients for optimization. In addition, this type of attack does not successfully recover the training set of DNNs.

\subsection{Generative Model Inversion Attacks}
To address the limitations that the recovered results are often non-semantic or meaningless when facing DNNs, Zhang et al. \cite{zhang2020secret} first proposed to train GAN\cite{goodfellow2020generative } using fuzzy or incomplete training set data to generate target label samples. Chen et al. \cite{chen2021knowledge} proposed to improve the training of GAN using public auxiliary data, and with the help of soft labels predicted by the target model. Yuan et al. \cite{yuan2023pseudo} further proposed to use of the target model to provide pseudo-labels and guide the training of conditional GAN\cite{mirza2014conditional} to optimize the latent space and decouple the generated data. However, the above methods are all white-box attacks and therefore may not be practical for real-world scenarios. 

To address this issue, Yang et al. \cite{yang2019neural} resorted to an auto-encoder (AE)\cite{baldi2012autoencoders} architecture to train an inversion model as a decoder to reconstruct the input data and perform data encoding with the target model as the encoder. In addition to these approaches, Zhu et al. \cite{zhu2022label} and Kahla et al. \cite{kahla2022label} investigate the direction of label-only model inversion attacks. Zhu et al. used the target model error rate to estimate the confidence of predictions and trained an attack model based on the framework of Yang et al. While Kahla et al. do not adopt the architecture of AE, train a GAN using a public dataset and propose a gradient estimation algorithm to estimate the true gradient, and optimize the potential input vector. Similar research only conducted for optimization algorithms continues to evolve. For example, Dionysiou et al. \cite{dionysiou2023exploring} proposed the use of evolutionary algorithms\cite{beyer2002evolution}, and Han et al. \cite{han2023reinforcement} used reinforcement learning\cite{mnih2015human} to optimize for input noise. However, the attack model in these related works is nothing more than the inversion model \cite{yang2019neural} or the same GAN \cite{zhang2020secret}\cite{kahla2022label}\cite{han2023reinforcement}. 

Currently, there is no powerful attack model other than the inversion model that can be trained using the target model in the label-only scenario. To the best of our knowledge, we are the first to investigate the practicable and powerful attack model in the label-only scenario.

\vspace{-0.25cm}

\subsection{Diffusion Models}
Diffusion Models (DMs) were first introduced in 2015 by Sohl-Dickstein et al. \cite{sohl2015deep}.  These models are inspired by non-equilibrium statistical physics \cite{jarzynski1997equilibrium} and work by systematically destroying the structure in a data distribution through an iterative forward diffusion process. The model then learns the reverse diffusion process to restore the data structure, resulting in highly flexible and easy-to-handle data generation models. In 2020, Ho et al. \cite{ho2020denoising} simplified this approach by proposing Denoising Diffusion Probabilistic Models (DDPMs). These models use parameterized Markov chains\cite{norris1998markov} trained by variational inference to generate samples matching the data after a finite amount of time. Dhariwal et al. \cite{dhariwal2021diffusion} later built on this work by proposing that prediction of noise through specific classifiers can guide sample generation. They demonstrated through systematic experiments that DM outperforms GAN. Ho et al.  \cite{ho2022classifier} then proposed “classifier-free diffusion guidance”, which aims to jointly train a conditional and unconditional diffusion model. The conditional diffusion model (CDM) is guided by the true labels, and the resulting conditional and unconditional diffusion models are combined to achieve a trade-off between sample quality and diversity. This is similar to the results obtained using classifier guidance.

Zheng et al. \cite{zheng2023targeted} first used pre-trained diffusion models for MIA, and they adopted the idea of previous white-box attacks, i.e., minimizing the cross-entropy loss between the target model prediction of the generated image and the true label, to optimize the $x_t$ in the noise reduction path. However, the final result of the attack makes the generation results unsatisfactory even though the classification accuracy is guaranteed. Therefore, a complete evaluation system is essential for MIA.

Combining label-only MIA and CDM, it was found that diffusion models guided by labels can be well applied to attack models in label-only MIA. However, unlike traditional diffusion models, the attacker will not have access to real training data and labels. To better reflect the knowledge behind the target model, an attack method was designed in Section IV.

\begin{figure*}[t]
  \centering
  \includegraphics[width=0.8\linewidth]{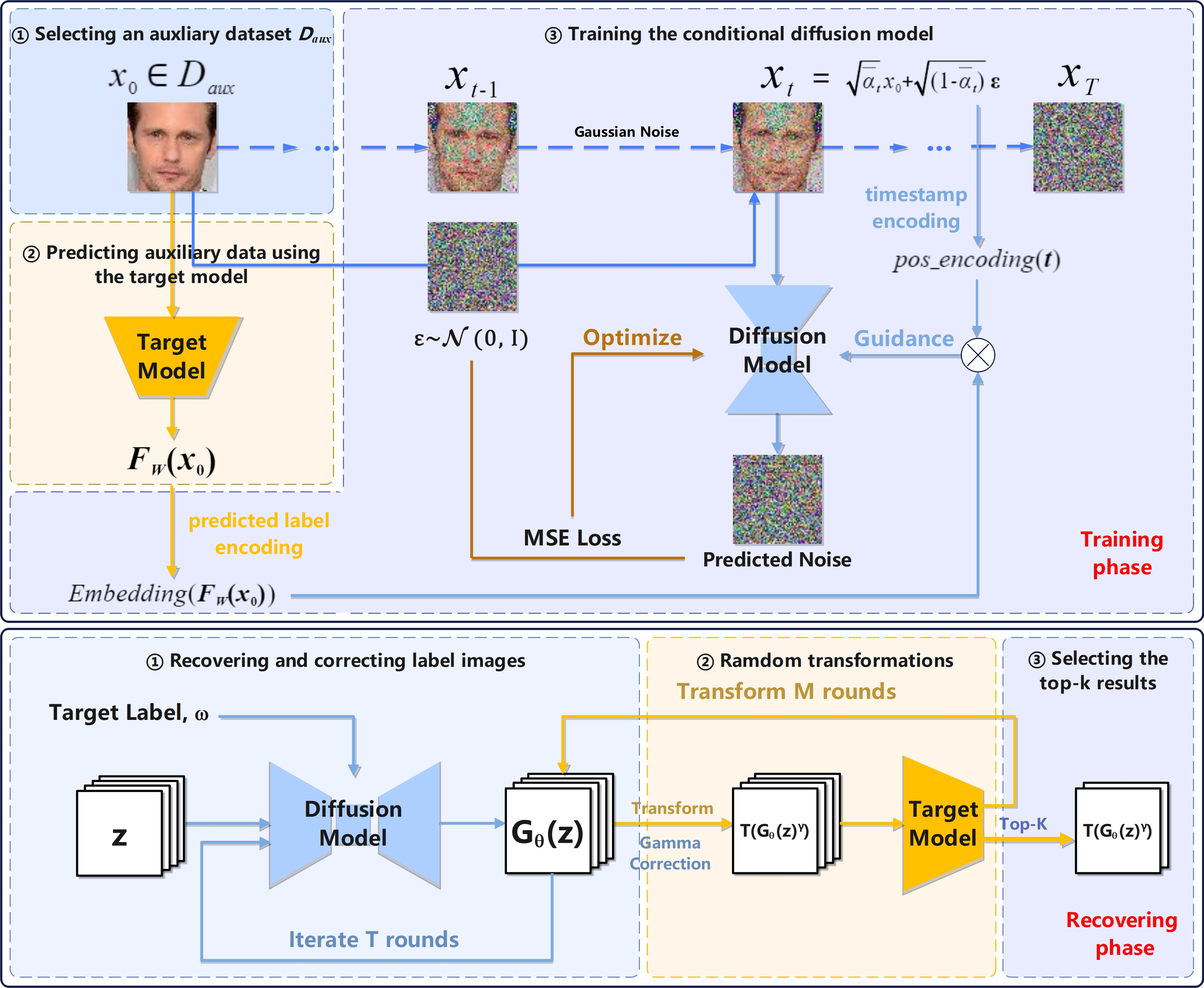}
  \caption{The attack overview of the proposed label-only model inversion attack method.}
  \label{figure2}
\end{figure*}

\section{Threat Model}
 In real attack scenarios, the target models are usually DNNs (e.g., face recognition models), and the sensitive data (face images) to be recovered often lie in a high-dimensional, continuous data space. However, recent works show that MIAs could even successfully reconstruct high-dimensional data, such as images \cite{fredrikson2015model}\cite{zhang2020secret}. In this study, we also adopt the convolutional neural network image classification model as our target model. Our focus is on the label-only scenario, where the adversary has access only to label predictions 
\begin{math}
  F_W (x)
\end{math}, obtained by inputting an image \begin{math}
  x
\end{math} into the target model
\begin{math}
  F_W
\end{math}.

{\bfseries Attack goal. }Given access to a target model 
\begin{math}
  F_{W}:[0,1]^{d}\rightarrow L
\end{math}, the attacker aimed to regenerate the representative samples \begin{math}
  \tilde{x}
\end{math} of the training dataset of the target label 
\begin{math}
  L
\end{math}; 
\begin{math}
  d
\end{math} represents the dimension of the input; 
\begin{math}
  L
\end{math} represents the predicted label, and “representative samples" means the generated images are similar to the target individual.

{\bfseries Task Knowledge. }We assume that the attacker knows the task of the target model, e.g., a face classification model. This assumption is reasonable as this information is typically available from the network or can be inferred through direct access to the target model.

{\bfseries Data Knowledge. }Based on the above assumptions, it is then reasonable to assume that the attacker can construct similarly distributed auxiliary datasets $D_{aux}$. Previous work has assumed that there is no target class overlap between the two datasets, but we argue that the possibility of target class overlap in the image recognition domain exists if an attacker can combine a large amount of auxiliary data for an attack. We discuss the impact of this scenario on the attack results in Section VI, but we weaken the assumption that the attacker does not know how many classes or images overlap, i.e., the attacker does not know which classes are overlapped.

\section{ATTACK METHOD DESIGN}
\subsection{Overview of Our Method}
This section details the design of our proposed method. As illustrated in Figure \ref{figure2}, our approach comprises two primary phases: the training phase and the recovery phase.

During the training phase, we train a generator for model inversion attacks. This involves the following steps: 
\begin{itemize}
    \item {\bfseries Step 1: }Selecting an auxiliary dataset $D_{aux}$ that is relevant to the target model task.

    \item {\bfseries Step 2: }Inputting  $\mathbf{x}_{0}\in D_{aux}$ into the target model $F_{W}$ yields the predicted labels $F_{W} \left ( \textbf{x}_{0}\right )$.
    
    \item {\bfseries Step 3: }Training a conditional diffusion model $G_{\theta }$ for the attack using the auxiliary dataset $D_{aux}$ from step 1 and employing the prediction labels $F_{W} \left ( \mathbf{x}_{0}\right )$ from step 2 as conditions to guide training. 
\end{itemize}

Our method improves existing MIA by using CDM as the attack model which is tailored for label-only scenarios. Instead of traditional CDM training, we use the target model’s predicted label for auxiliary data as guidance. This aligns the auxiliary data under the predicted label with the target model’s decision for the target, helping the attack model learn consistent features of the target from the data.

During the recovery phase, we use the trained generator to recover target label data. This involves the following steps:
\begin{itemize}
    \item {\bfseries Step 1: }Inputting multiple standards normally distributed noise images \begin{math}
    \mathbf{z}
\end{math} and target attack label \begin{math}
    l
\end{math} into the trained conditional diffusion model to recover images \begin{math}
    G_{\theta }\left ( \mathbf{z}\right )
\end{math} of this label with a pre-defined guidance strength. Unlike traditional CDM, there is a quality impact on the generated data since the data under identical guidance is not the same entity. To solve this issue, we use gamma correction to correct the generated images to $G_{\theta }\left ( \mathbf{z}\right )^{\gamma }$.

\item {\bfseries Step 2: }Transforming the corrected generated image \begin{math}
    G_{\theta }\left ( \mathbf{z}\right )^{\gamma }
\end{math} randomly into \begin{math}
    T\left (G_{\theta }\left ( \mathbf{z}\right )^{\gamma } \right )
\end{math} and inputting it into the target model for prediction, then repeating this step \begin{math}
    M
\end{math} times.

\item {\bfseries Step 3: }Selecting the top-$k$ robust generated images from  $T\left (G_{\theta }\left ( \mathbf{z}\right )^{\gamma } \right )$, i.e., the top-$k$ images that still predict the target label with the highest ratio among \begin{math}
    M
\end{math} random changes. In contrast to the low fault-tolerance limitation of existing attack models, which can only generate a single image for the target, our approach is not only powerful but also has unique advantages.

\end{itemize}

\vspace{-0.32cm}

\subsection{The Training Phase}

This section presents a detailed analysis of each step in the training phase. 

\subsubsection{Selecting an auxiliary dataset}

To achieve better generator results, it is essential to choose an auxiliary dataset that closely resembles the training set in terms of data distribution and task relevance. Previous research \cite{struppek2022plug} has demonstrated that the auxiliary dataset significantly affects the results of the generator. For instance, if the target model is for face recognition but the generator is trained on an oil painting face dataset, the inversion attack results will produce unsatisfactory oil painting images of the target face. Additionally, we opted for a larger-scale public dataset and preprocess to extract a considerable amount of feature information. The generated results can approximate private ones by learning from these public features. For instance, for face image recovery, the auxiliary data should include only facial regions to avoid background influence and enable direct learning by the generator.

\subsubsection{Predicting auxiliary data using the target model}
In previous black-box attacks, the assistance of the target model in training a powerful generator was often overlooked due to technical and scenario limitations. As a result, these researches focused on optimization strategies. The conditional diffusion model is a type of generator that does not require back-propagation of gradient optimization after loss calculation through label and prediction. However, real training data and labels cannot be obtained. Therefore, auxiliary data ${x}_{0}\in D_{aux}$ selected in the first step is fed into the target model $F_{W}$ for prediction, to obtain the target model’s classification task labels $ F_{W} \left ( {x}_{0}\right )$. This reflects the target model’s judgment on the feature of the training data. For a simple instance, if the target model is a “0-4” handwritten digit classifier, then input digits “7, 9” will have a high probability of outputting “1”. Thus, we can learn “1” by “7, 9”.

% Previous black-box attacks have neglected the help of the target model in generator training due to technical and scenario limitations to optimize a powerful generator, leading to research focusing on optimization strategies. To facilitate data recovery during the recovery phase without the need for optimization, we employ predicted label-guided training. Since black-box attacks cannot be optimized by computing gradients in the recovery phase to generate target label images as white-box do, we leverage the target model \begin{math}
%     F_{W}
% \end{math} to predict and assign a label \begin{math}
%     F_{W} \left ( {x}_{0}\right )
% \end{math} to each auxiliary data \begin{math}
%     {x}_{0}\in D_{aux}
% \end{math}, and then we use \begin{math}
%     F_{W} \left ( {x}_{0}\right )
% \end{math} as a condition to guide the generator, enabling it to target specific labels for noise prediction.

\subsubsection{Training the conditional diffusion model}
We train the conditional diffusion model \begin{math}
    G_{\theta }
\end{math} using the data and labels provided in the preceding two steps. In reference to previous work \cite{ho2020denoising}\cite{ho2022classifier}, the training procedure for the conditional diffusion model comprises two primary stages: forward diffusion and backward prediction. During forward diffusion, Gaussian noise is added to the auxiliary data \begin{math}
    x_{0}
\end{math} T times, ultimately resulting in standard normally distributed noise. At each step, Gaussian noise is added to the data \begin{math}
    x_{t-1}
\end{math} obtained in the previous step as follows:
\begin{equation}
  q\left ( x_{t}\mid x_{t-1} \right )=\mathcal{N}\left ( x_{t};\sqrt{1-\beta _{t}}x_{t-1},\beta _{t}\mathrm{\mathbf{I}}\right )
\end{equation}

Where \begin{math}
  \beta_t  
\end{math} represents the noise variance at each step, ranging from 0 to 1. A linear variance schedule is employed to customize the variance at each step, resulting in \begin{math}
    x_t
\end{math} being generated at each step of the diffusion process and the entire process is fixed as a Markov chain, as shown below:
\begin{equation}
  q\left ( x_{1:\textrm{T}}\mid x_{0}\right ) = \prod_{t=1}^{\textrm{T}}q\left ( x_{t}\mid x_{t-1} \right )
\end{equation}

Given this property, we only need to sample and train the \begin{math}
    t_\textrm{th}
\end{math} step of the training process. Based on the first two formulas, we derive the following equation:
\begin{equation}
  q\left(\left.x_t\right|x_0\right)=\mathcal{N}\left ( x_{t};\sqrt{{\bar{\alpha}}_t}x_0,(1-{\bar{\alpha}}_t)\mathrm{\mathbf{I}}\right )
\end{equation}

Where \begin{math}
  \alpha _{t}=1-\beta _{t} 
\end{math}, and \begin{math}
    {\bar{\alpha}}_t=\prod_{i=1}^{t}\alpha_i
\end{math}, thus \begin{math}
    x_t=\sqrt{{\bar{\alpha}}_t}x_0+\sqrt{(1-{\bar{\alpha}}_t)}\varepsilon, \: \varepsilon\sim\mathcal{N}\left ( \mathbf{0},\mathrm{\mathbf{I}}\right )
\end{math}. This means that the noise data \begin{math}
    x_t
\end{math} obtained after adding $t$ times of noise can be directly derived from the original auxiliary data \begin{math}
    x_0
\end{math}. Since the variance increases linearly, the limit of \begin{math}
    {\bar{\alpha}}_t
\end{math} approaches 0 as long as T becomes sufficiently large, resulting in \begin{math}
    x_\textrm{T}
\end{math} being close to standard normally distributed noise.

Backward prediction involves predicting the noise added during forward diffusion. This is accomplished using a U-Net \cite{ho2020denoising} neural network model comprising downsampling blocks, upsampling blocks, and attention blocks. The input is \begin{math}
    x_t
\end{math}, the added noise is predicted and optimized using \begin{math}
    D_{KL}( q\left ( x_{t-1}\mid x_{t},x_{0} \right )\parallel p_{\theta }( x_{t-1}\mid x_{t}))
\end{math}, where \begin{math}
    p_{\theta }( x_{t-1}\mid x_{t})
\end{math} represents the predicted noise distribution and \begin{math}
    q\left ( x_{t-1}\mid x_{t},x_{0} \right)
\end{math} denotes the real posterior distribution. Denoising diffusion probabilistic models (DDPM) experiments \cite{ho2020denoising} have demonstrated that calculating the Mean Squared Error (MSE) loss between the predicted noise and random Gaussian noise yields better optimization results.  Thus, during the training phase, we only need to calculate \begin{math}
    \bigtriangledown _{\theta }\left \| \varepsilon -\varepsilon _{\theta }(x_{t},t)\right \|^{2}
\end{math}where \begin{math}
    \varepsilon
\end{math} represents the Gaussian noise added from step 0 to step $t$, and the \begin{math}
    \varepsilon_\theta(x_t,t)
\end{math} denotes the noise predicted by the generator \begin{math}
    G_\theta 
\end{math} based on \begin{math}
    x_t,t
\end{math}. 

To ensure that the training process is guided by the predicted labels, we incorporate labels encoded via the Embedding function of PyTorch into each timestamp. The timestamp $t$ is initially encoded via the position encoding function \cite{NIPS2017_3f5ee243} to a fixed dimension, and then the label is encoded into the same dimension by the Embedding function, which in turn sums the two. This is followed by a time embedding layer at each sampling block of the diffusion model, which is synchronized with the hidden layer feature dimension of the current block through the repeat function. The time embedding layer comprises a SiLU activation layer and a Linear layer. The specific semantics of this approach is to enable the diffusion model to learn the noise distribution added to the predicted target $F_W(x_0)$. Therefore, noise reduction can be carried out during the recovery phase by progressively embedding the target labels into the timestamps. This process ultimately results in the generation of outcomes specific to that target.

To ensure the training of the conditional diffusion model does not over-fit the label information, a certain probability $p$ is introduced. This allows the training to optimize $G_\theta$ without the need for label guidance, i.e., $\bigtriangledown _{\theta }\left \| \varepsilon -\varepsilon _{\theta }[x_t,(\textit{pos\_encoding}(t)+\textit{Embedding}(F_W(x_0)))]\right \|^{2}$ or $\bigtriangledown _{\theta }\left \| \varepsilon -\varepsilon _{\theta }(x_{t},t)\right \|^{2}$. Consequently, the attack model learns not only from the labels but also from the inherent features and structure of the data. This approach improves the generalization capability of the attack model, thereby ensuring the diversity and authenticity of the generated data. The method for determining the value of probability $p$ is elaborated in Section VI.

To summarize, during the training process, we initially assign each auxiliary data a random timestamp $t$ that falls between 1 and T. Following this, $x_t$ can be computed from Equation (3) after iterative noise addition from $x_0$ after $t$ steps. Subsequently, the encoded predicted label of $x_0$ is embedded as the condition into the timestamp. Finally, the MSE loss between the true Gaussian noise in Equation (3) and the noise predicted by the conditional diffusion model is calculated. This is iteratively optimized, with the optimization objective shown below:
\begin{equation}
 \begin{gathered}
   \textrm{min}\:\:\mathcal{L}_{\textrm{MSE}}\left (\varepsilon ,\varepsilon_\theta(x_t,t,F_W(x_0)) \right ) = \\ \left \| \varepsilon -\varepsilon_{\theta }[x_t,(\textit{pos\_encoding}(t)+\textit{Embedding}(F_W(x_0)))] \right \|^{2}
  \\\textrm{s.t.}\:\: \Pr [F_W(x_0)=\varnothing]=p  
 \end{gathered}
 \centering
\end{equation}

\subsection{The Recovering Phase}
This section describes how we recover data for corresponding target labels using the conditional diffusion model trained in Section IV.B. 

\subsubsection{Recovering and correcting target label images}
We input target label \begin{math}
    l
\end{math}, guidance strength $\omega$ and standard normally distributed noise image \begin{math}
    \mathbf{z}
\end{math} into the trained U-Net to predict the noise and gradually denoise over T rounds. In order to sample \begin{math}
    x_{t-1}\sim p_{\theta}(x_{t-1} \mid x_t)
\end{math}, based on the above available information it is only necessary to calculate:
\begin{equation}
\begin{aligned}
    x_{t-1}=\frac{1}{\sqrt{{\bar{\alpha}}_t}}\left(x_t-\frac{\beta_t}{\sqrt{\left(1-{\bar{\alpha}}_t\right)}}\varepsilon_\theta\left(x_t,t\right)\right)+\sqrt{\beta_t}\mathbf{z} \\,  \textrm{where}\:\:\mathbf{z}\sim \mathcal{N}(\mathbf{0},\mathcal{\textbf{I}})
\end{aligned}
\end{equation}

\begin{figure}[t]
  \centering
  \includegraphics[width=0.62\linewidth]{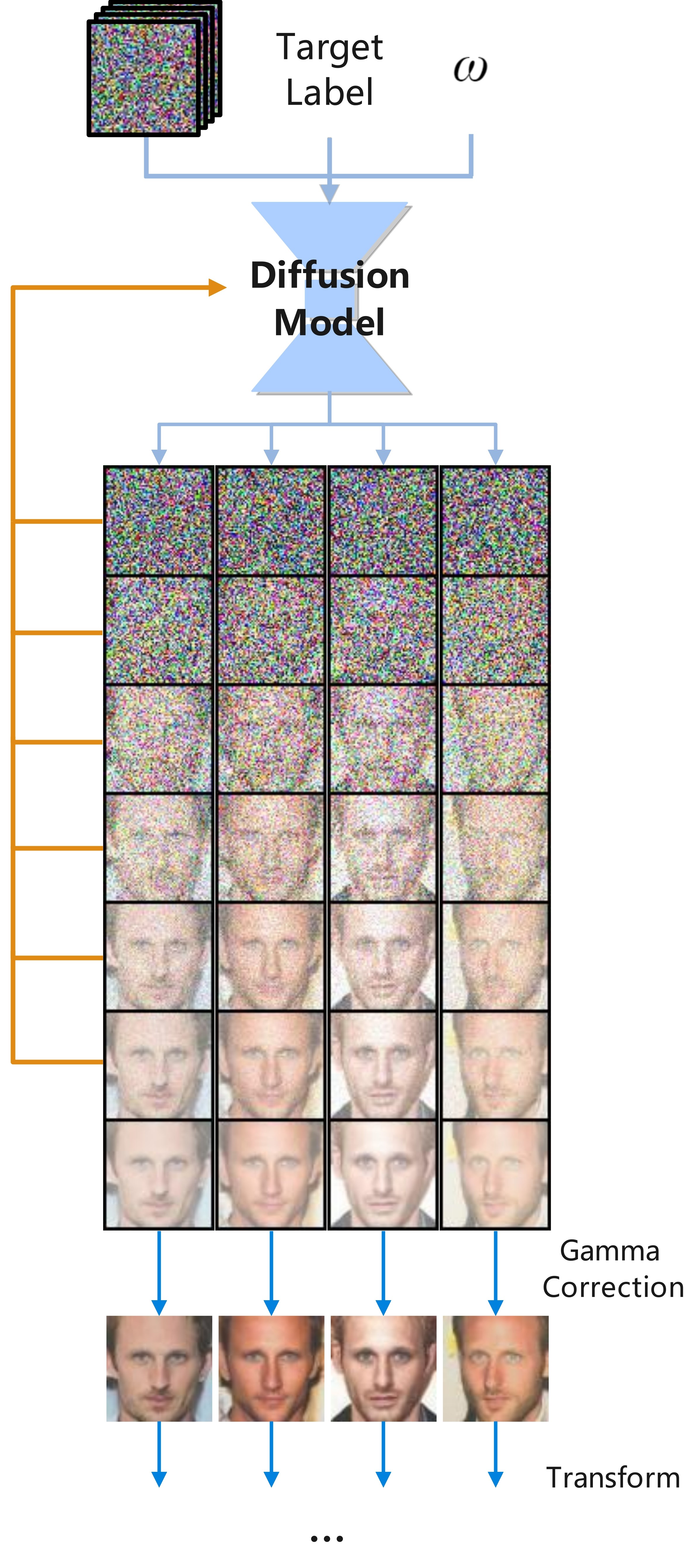}
  \caption{The first step in the recovery phase is to input noise, target labels, and guidance strength $\omega$ to the trained diffusion model and denoise them step by step to obtain the generated image, and eventually correct it.}
  \label{figure3}
\end{figure}
To use a target label to guide recovery, according to \cite{ho2022classifier}, we need to modify the predicted noise in Eq. (5) as follows:
\begin{equation}
\begin{gathered}
{\varepsilon_\theta\left(x_t,t\right)\rightarrow\widetilde{\varepsilon}}_\theta\left(x_t,t,l\right)= \\ \left(1+\omega\right)\varepsilon_\theta[x_t,(\textit{pos\_encoding}(t)+\textit{Embedding}(l))]-\omega\varepsilon_{\theta}(x_t,t)
\end{gathered}
\end{equation}    
The semantics of doing so in Eqs. 5,6 is precisely the step-by-step noise prediction based on the input target $l$ and the noise reduction accordingly, with the guided training in the training phase echoing. Where $\omega$ represents the strength of the guidance provided by the target labels. As can be learned from Equation 6, if the strength is higher, then the proportion of conditional prediction noise is greater. This results in the generated image features being closer to the target, but at the expense of the quality of the generation.

After T rounds of denoising, we obtain the representative recovered images \begin{math}
   G_\theta(\mathbf{z}) 
\end{math} of the target label. However, since diffusion models vary in the samples of data that may be learned for specific labels, this leads to a gap between the results of noise reduction and those generated by traditional diffusion models. However, the purpose of essentially learning the added noise distribution for a specific target is achieved, so it is only necessary to do a secondary correction for the generated image. 

We apply gamma correction \cite{hunt2005reproduction} to the generated image, i.e., \begin{math}
  G_\theta(\mathbf{z})\rightarrow{A \cdot G_\theta(\mathbf{z})}^\gamma  
\end{math}, adjusting it to match the human eye’s perception. Where \begin{math}
   \gamma 
\end{math} is the gamma factor, and the value of $A$ is usually 1. The graphical representation of the recovery process can be observed in Figure \ref{figure3}.  According to \cite{hunt2005reproduction}, $\gamma$ = 2.2 aligns more closely with the human eye’s judgment of brightness and color. As such, the value of $\gamma$ can be set to approximately 2.2. Section VI.C.4 provides a comparison of the specific impact of different gamma values on the results, and shows that $\gamma$ = 2.2 is indeed more consistent with human perceptual similarity judgments.

In addition, since the number of recovered images can be large, we need to filter out the most representative $k$ images. In a black-box scenario, our strategy involves performing multiple random transformations and making predictions to select the most robust generated image. 

\subsubsection{Random transformations}
Thus, we randomly transform ${G_\theta(\mathbf{z})}^\gamma$ by randomly cropping and flipping it vertically or horizontally with a certain probability. The transformed image \begin{math}
    {T(G_\theta\left(\mathbf{z})\right)}^\gamma
\end{math} is then fed into the target model for prediction, and this process is repeated \begin{math}
  M  
\end{math} times.

\subsubsection{Selecting the top-k robust generated images}
We calculate the representative weights \begin{math}
    \mathbb{E}[\delta (F_W(T(G_\theta(\mathbf{z})^{\gamma})),l)]
\end{math} for each image based on the above information, as follows:
\begin{equation}
    \mathbb{E}[\delta (F_W(T(G_\theta(\mathbf{z})^{\gamma})),l)]=\frac{1}{M}\sum_{i=1}^{M}\delta (F_W(T_i(G_\theta(\mathbf{z})^{\gamma})),l)
\end{equation}

where the \begin{math}
   \delta 
\end{math} function returns 1 when its two input values are equal and 0 when they are different. The reason why the $\delta$ function compares the predicted labels with the target labels is that the adversaries can only obtain the label information by accessing the target model $F_{W}$. Further, using the above equation, we select the \begin{math}
   G_\theta(\mathbf{z}) 
\end{math} corresponding to the top-$k$ largest representative weights.

\begin{figure}[t]
  \centering
  \includegraphics[width=0.89\linewidth]{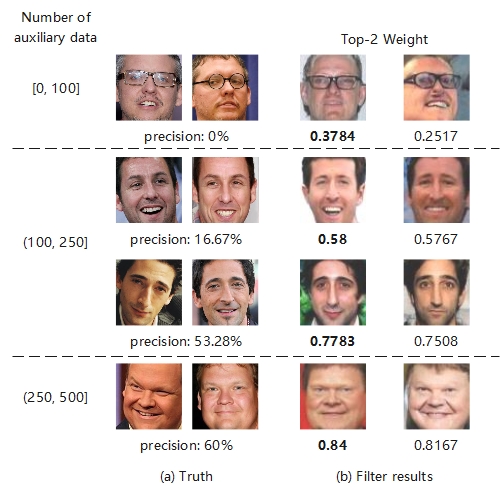}
  \caption{The impact of variations in the target model’s precision rate for classifying different individuals on weight filtering is depicted. (a) represents the true images of an individual, with the target model’s test precision rate for each individual. (b) represents the two most optimal results after filtering according to the weights  \begin{math}
    \mathbb{E}[\delta (F_W(T(G_\theta(\mathbf{z})^{\gamma})),l)]
\end{math}.}
  \label{additional}
\end{figure}

It’s crucial to highlight that the target model exhibits varying classification precision for different individuals, which significantly influences the generation of results and the calculation of weights \begin{math}
    \mathbb{E}[\delta (F_W(T(G_\theta(\mathbf{z})^{\gamma})),l)]
\end{math}. As depicted in Figure \ref{additional}, the individuals in (a) display increasing classification precision from top to bottom. To verify the classification robustness for the corresponding individuals, we also performed 100 random transformations in step 2 for the test set and calculated the average precision, which is 13.5\%, 25\%, 54\%, and 58.7\% for the four individuals on the graph, respectively. The increase in precision and robustness also provides the target individuals with more auxiliary data that can be used to train the attack model, as shown in the data interval on the left side of Figure \ref{additional}(a). Because the target model that more accurately extracts the individual's features will assign more auxiliary data that are close to the target.

From the results in (b), it can be observed that when the target model is more accurate in classifying that target and the more data can be used for training, the higher the calculated weights \begin{math}
    \mathbb{E}[\delta (F_W(T(G_\theta(\mathbf{z})^{\gamma})),l)]
\end{math} are, i.e., the generated images are classified under the target class after many random transformations. For example, the first individual has a test set precision of 0\%, and the maximum weight calculated for the generated data according to Equation 7 is 0.3784, which means that the rest of the generated data are all below this value. The rest of the individuals have gradually increased weight values as the accuracy rate and robustness increase, and the gap between the maximum weight and the second largest weight is narrowed. This indicates that the quality of the generated data has also been improved, and this characteristic can be identified through qualitative judgment.

\section{Discussion of the attack method}
\subsection{The Training Phase}
While the first and second steps of the training phase serve as a foundation for the subsequent conditional diffusion model training, each stage holds its significance. For instance, the proximity of the auxiliary dataset to the training set and the number of images that can be assigned to each label within it can influence the results (see Section VI.C.3 for details). Moreover, it is intuitively preferable to assign target labels to data that exhibit extreme closeness to the features. Apart from the impact of the dataset, labels play a pivotal role in our approach by guiding the diffusion model toward making noisy predictions for specific targets. The third step constitutes the crux of the training phase and through this process, we can summarize the following advantages of the conditional diffusion model:
\begin{enumerate}
    \item {\bfseries Applied to label-only black-box scenarios. }The state-of-the-art approach \cite{yuan2023pseudo} using conditional GAN requires the computation of gradients by the target model for conditional guidance, but our training process only requires the encoding of predicted labels with embedded timestamps. Therefore, it is not necessary to obtain gradient information with the help of a target model to train a powerful attack model in a label-only scenario. In contrast to white-box attacks, the effectiveness of an attack model is not directly impacted by the architecture and parameters of the target model. Instead, it is determined by the target model’s ability to accurately judge the features of the auxiliary data. As such, the more accurately the target model can judge these features, the more effective the attack model training will be. This proposition is verified in Section VI.C.
    
    \item {\bfseries Stable training process. }In contrast to conditional GAN or GAN, training conditional diffusion models does not require adversarial or sophisticated loss functions. As shown in Equation 4, our training uses a uniform loss function to narrow the \begin{math}
    D_{KL}( q\left ( x_{t-1}\mid x_{t},x_{0} \right )\parallel p_{\theta }( x_{t-1}\mid x_{t}))
\end{math}. 

    \item {\bfseries Avoiding MIA overfitting. }As opposed to normal training, MIA requires training the generator on auxiliary data to enable the recovery of the target training set data. The GAN itself suffers from the mode collapse problem \cite{bau2019seeing}, and training the generator on the auxiliary data set results in a worse fit. However, the conditional diffusion model is trained for specific labels to learn how to recover the probability distribution of the original data from the noise.
\end{enumerate}

\subsection{The Recovering Phase}
Unlike previous black-box attacks \cite{kahla2022label}\cite{dionysiou2023exploring}\cite{yoshimura2021model}\cite{han2023reinforcement}, our recovery phase does not employ any optimization algorithms to emphasize the power of the generator itself. Instead, it relies on sampling and filtering to demonstrate that the generator’s ability can directly determine the generated results. 

% Furthermore, the generated results are also influenced by the guiding strength $\omega$. For better access to target features, a larger strength can be set. Conversely, for a trade-off between realism and feature ability, an appropriate strength can be chosen (refer to Section VI.C for further details). 

Lastly, through the recovery process, the following advantages of the conditional diffusion model can be summarized:
\begin{enumerate}
    \item {\bfseries Generate multiple results for the same label. }The Inversion model \cite{yang2019neural}\cite{zhu2022label} of the AE architecture can only generate unique results for a specific label, but the conditional diffusion model can generate a diversity of target results based on random input noise. And without optimization, only the conditional diffusion model can be attacked for a specific label.
    
    \item {\bfseries Targeted attack without optimization algorithm. }The conditional diffusion model can generate the corresponding target by adjusting the $\omega$ without an optimization algorithm. Low strength means that the recovery result may not be close to the target but the image is more realistic, while high strength means that the target features are recovered to the maximum extent even though the image is not realistic. And with the help of the parameter $\omega$ and the special generation process of the diffusion model, we believe that future black-box optimization strategies can do better on this basis.
\end{enumerate}

\section{Experiments}

In this section, we will evaluate this method not only quantitatively with relevant metrics, but also qualitatively from the perspective of visual inspection to assess the authenticity and compare it with the effect of generators in related work. The specific experimental setup and evaluation metrics are shown below.

\vspace{-0.32cm}
\subsection{Experimental Setup}

\subsubsection{Datasets}
Our experiments were conducted on two face recognition datasets and one handwritten digits dataset, and the detailed data allocation is shown in Table \ref{table3}: 
\begin{itemize}
\item {\bfseries FaceScrub \cite{ng2014data}. }FaceScrub is a URL dataset with 100,000 images of 530 actors, which contains 265 male actors and 265 female actors. However, since not every URL was available during the writing period, we downloaded a total of 43,149 images for 530 individuals and resized the images to 64 × 64.

\item {\bfseries CelebA \cite{liu2015deep}. }CelebA is a dataset with 202,599 images of 10,177 celebrities from the Internet. We used the same crop as \cite{yang2019neural}\cite{zhu2022label} to remove the background of images in this dataset other than faces to reduce the impact on the experiment. There are 296 individuals overlapping in the two datasets, and since we need to discuss the impact of whether there are individuals overlapping in the two datasets on the experiment, we removed a total of 6,878 images of 296 individuals from CelebA and similarly resized the images to 64 × 64.

\item {\bfseries MNIST \cite{lecun1998gradient}. }A dataset composed of 70,000 handwritten digit images in 10 classes. Each image is resized to 64 × 64.
\end{itemize}

\subsubsection{Target Model}
We adopt the same target model architecture as \cite{yang2019neural}, which consists of four CNN blocks followed by two fully connected layers. Each CNN block contains a convolutional layer, a batch normalization layer, a max-pooling layer, and a ReLU activation layer. We train this target model on the FaceScrub dataset using the same hyperparameters as \cite{yang2019neural} and achieve a test set accuracy of 83.82\% for face recognition. The model outputs the predicted confidence scores for the input face image, i.e., the probabilities of belonging to each of the 530 possible individuals. However, based on our assumptions in Section 3, we modify the model output to predict the label instead of the confidence scores. Note that we use numeric labels to represent the individuals, which does not affect the experimental results since we only need a single number for the label-only output.

In addition to this, in order to argue the impact of the target model capability for this attack, we used different CNNs as follows: (1) VGG16 adapted from \cite{simonyan2014very}; (2) ResNet-152 adapted from \cite{hek2016}.

\begin{table}[t]
\centering
  \caption{Data allocation of the classifier and its attack model}
  \label{table3}
  \setlength{\tabcolsep}{1.32mm}{
\begin{tabular}{@{}cc|c@{}}
\toprule
\multicolumn{2}{c|}{Classifier}                                                                                                                                                         & Attack Model                                                                              \\ \midrule
\multicolumn{1}{c|}{Task}                                                                               & Data                                                                          & Auxiliary Data                                                                            \\ \midrule
\multicolumn{1}{c|}{\multirow{2}{*}{\begin{tabular}[c]{@{}c@{}}FaceScrub\\ (530 classes)\end{tabular}}} & 80\% train, 20\% test                                                         & \begin{tabular}[c]{@{}c@{}}\textbf{Overlap:} CelebA\\ (296 individuals overlapping)\end{tabular}   \\ \cmidrule(l){2-3} 
\multicolumn{1}{c|}{}                                                                                   & 80\% train, 20\% test                                                         & \begin{tabular}[c]{@{}c@{}}\textbf{Nonoverlap:} CelebA\\ (non-individuals overlapping)\end{tabular}   \\ \midrule
\multicolumn{1}{c|}{\multirow{2}{*}{\begin{tabular}[c]{@{}c@{}}MNIST\\ (10 classes)\end{tabular}}}      & 50\%train, 50\% test                                                          & \textbf{Overlap:} MNIST 20\% test data                                                             \\ \cmidrule(l){2-3} 
\multicolumn{1}{c|}{}                                                                                   & \begin{tabular}[c]{@{}c@{}}80\% train, 20\% test\\ (labels: 0-4)\end{tabular} & \begin{tabular}[c]{@{}c@{}}\textbf{Nonoverlap:} MNIST’s other 5 labels\\ (label: 5-9)\end{tabular} \\ \bottomrule
\end{tabular}}
\end{table}

\subsubsection{Attack Model and Implementation Details}
The attack model used in our method is U-Net, which is composed of a double convolution block, 3 downsampling blocks, 2 bottom double convolution blocks, 3 upsampling blocks, and a convolution layer, where the double convolution block is composed of 2 convolution layers, 2 group normalization layers, and a GELU activation layers in the order of activation layer in the middle and two layers on each side. The downsampling block consists of a max-pooling layer, two convolutional layers, a SiLU activation layer, a linear connection layer, and a self-attentive layer, while the upsampling block differs in that the max-pooling layer is replaced by an inverse convolutional layer. The U-Net is trained for a maximum of 300 iterations, where CelebA is the training set, the batch size is 16, the learning rate is 3e-4 and the last 50 iterations are reduced to 1e-4. Secondly, the MSE loss function, AdamW optimization algorithm with Exponential Moving Average mechanism is used. In addition, the noise step of forward diffusion is 1500 and the variance schedule is linear, where \begin{math}
    \beta_0=1e-4, \beta_t=0.02
\end{math}. Moreover, the gamma factor \begin{math}
    \gamma
\end{math} of 2.3, the guidance strength \begin{math}
    \omega
\end{math} of 4, and the probability $p$ of 0.1. Given that the image dimensions of the auxiliary datasets are uniformly 64 × 64, both training phases necessitate 28G of graphics memory, while both recovery phases require 26G of graphics memory. The batch size and the number of parameters in the target model directly influence the memory size in a proportional relationship. Furthermore, the number of auxiliary datasets directly affects the training time. Utilizing CelebA as the auxiliary dataset, we employ two NVIDIA RTX 3090 GPUs and Inter Xeon Platinum 8350C CPUs to complete a training round in an average duration of 33 minutes, with the generation of 48 images taking an average of 4 minutes. When employing MNIST as the auxiliary dataset, a training round with identical equipment takes an average of 5 minutes, with the total time for training and sampling amounting to 1.5 days.

\subsubsection{Comparison of Attack Methods}

Considering that none of the current generators in black-box MIA can do what the methods in this paper do in the FaceScrub task, we selected two white-box attack methods and one black-box attack method as baselines for fair comparison: Generative Model Inversion (\textbf{GMI}) \cite{zhang2020secret}, Pseudo Label-Guided Model Inversion Attack (\textbf{PLG}) \cite{yuan2023pseudo}, and Learning-Based Model Inversion (\textbf{LB-MIA}) \cite{yang2019neural}. Due to the low complexity of handwritten digital images (a single channel and most of the pixel values are 0), LB-MIA can do equally well, so for the MNIST task, only comparisons are made with this. All three methods employ a generator for attack purposes, with GMI and PLG utilizing a generator from GAN and LB-MI employing an inversion model from an AE framework. GMI necessitates gradient descent to generate representative data for the target label, thus we did not modify the original training and recovery stages. PLG can generate representative data by inputting the target label and latent vector, so we adopted the same selection strategy as our own method. However, the original strategy of assigning top-$n$ images to each individual proved unreliable under a label-only setting during the training stage. Consequently, we used all auxiliary data for training and trained under a white-box setting to facilitate a fairer comparison between our diffusion model and GAN trained under a white-box setting. Lastly, in order to enable LB-MIA to generate the most representative data for the target label, we directly encoded the label as a corresponding one-hot vector to train the inversion model. All methods, including our own, trained their respective attack models on the same auxiliary dataset and performed model inversion attacks on the same target model trained on the same training set.

\vspace{-0.32cm}
\subsection{Evaluation Metrics}

In this section, we conduct a comprehensive evaluation of the generated data in terms of its accuracy and similarity to target individuals, as well as its realism from a visual perspective. Our evaluation approach is more extensive and closely approximates that of a white-box attack relative to related black-box attacks. The specific evaluation metrics are detailed below.

{\bfseries Attack Accuracy (Attack Acc). }This metric is employed to quantitatively evaluate the ability of generated images to accurately identify target individuals. To this end, we trained an evaluation model with a distinct architecture from the target model and higher test accuracy to serve as a proxy for human judgment. Attack accuracy is determined by calculating the percentage of $k$ generated images classified as the target label and averaging the results over 530 individuals. The evaluation model, a ResNet-18 \cite{hek2016} trained using the same training set as the target model, achieved a test set accuracy of 93.03\%. The evaluation model accuracy and allocation in all tasks are shown in Table \ref{table4}.

{\bfseries K-Nearest Neighbor Distance (KNN Dist). }KNN Dist is the shortest feature distance from a reconstructed image to the real private training images for a given target. This metric is used to evaluate the similarity at the feature level. Furthermore, the feature distance is measured by the \begin{math}
    l_2
\end{math} distance between two images when projected onto the feature space, i.e., the output of the penultimate layer of the evaluation classifier. It is important to note that different evaluation models produce different feature dimensions, resulting in variations in the value of this metric across models. However, it is sufficient to compare the magnitude of the metric within the same evaluation model.

{\bfseries Frechet Inception Distance (FID). }FID \cite{heusel2017gans} is commonly used in the work of GAN to evaluate the generated images. FID score measures feature distances between real and fake images, and lower values indicate better image quality and diversity. However, there is inaccuracy in the evaluation of this metric due to the unsuitability of Inception v3 for extracting face features and the inaccuracy of FID calculations based on a small number of generated images \cite{zheng2023targeted}. In contrast to white-box attacks, black-box attacks basically do not use this metric.

{\bfseries Learned Perceptual Image Patch Similarity (LPIPS). }LPIPS \cite{zhang2018unreasonable} is a metric that measures the perceptual similarity between two images by calculating the similarity between activations of image blocks within a predefined network. This metric has been demonstrated to closely align with human perception and is employed in our experiment to evaluate perceptual similarity. AlexNet \cite{krizhevsky2012imagenet}, which serves as the default predefined network, performs optimally as a forward metric and closely approximates human perception. However, this does not imply that VGG is inferior in terms of matching human perception; when used for optimization, VGG more closely resembles traditional perceptual loss. Consequently, both networks are utilized in our evaluation.

\begin{table}[t]
\centering
  \caption{Allocation of the target model and evaluation model.}
  \label{table4}
  \setlength{\tabcolsep}{3mm}{
\begin{tabular}{@{}ccc|cc@{}}
\toprule
\multicolumn{3}{c|}{Target model}                                                                                                               & \multicolumn{2}{c}{Evaluation model}                                       \\ \midrule
\multicolumn{1}{c|}{Task}                       & \multicolumn{1}{c|}{Model}                                                         & Accuracy & \multicolumn{1}{c|}{Model}                      & Accuracy                 \\ \midrule
\multicolumn{1}{c|}{\multirow{3}{*}{FaceScrub}} & \multicolumn{1}{c|}{CNN}                                                           & 83.82\%  & \multicolumn{1}{c|}{\multirow{3}{*}{Resnet-18}} & \multirow{3}{*}{93.03\%} \\ \cmidrule(lr){2-3}
\multicolumn{1}{c|}{}                           & \multicolumn{1}{c|}{VGG16}                                                         & 88.22\%  & \multicolumn{1}{c|}{}                           &                          \\ \cmidrule(lr){2-3}
\multicolumn{1}{c|}{}                           & \multicolumn{1}{c|}{IR152}                                                         & 89.68\%  & \multicolumn{1}{c|}{}                           &                          \\ \midrule
\multicolumn{1}{c|}{\multirow{2}{*}{MNIST}}     & \multicolumn{1}{c|}{\begin{tabular}[c]{@{}c@{}}Resnet-18\\ (Overlap)\end{tabular}} & 99.31\%  & \multicolumn{1}{c|}{CNN}                        & 99.5\%                   \\ \cmidrule(l){2-5} 
\multicolumn{1}{c|}{}                           & \multicolumn{1}{c|}{\begin{tabular}[c]{@{}c@{}}CNN\\ (Nonoverlap)\end{tabular}}    & 99.86\%  & \multicolumn{1}{c|}{Resnet-18}                  & 99.94\%                  \\ \bottomrule
\end{tabular}}
\end{table}

\vspace{-0.32cm}
\subsection{Experimental Results}

\begin{table*}
  \centering
  \caption{Quantitative evaluation and attack performance comparison on various methods under whether FaceScrub and CelebA datasets overlap or not. ↑ and ↓ respectively symbolize that higher and lower scores give better attack performance.}
  \label{table5}
  \setlength{\tabcolsep}{3.6mm}{
  \begin{tabular}{clcccccccc}
    \toprule
    \multicolumn{2}{c}{Dataset}                       &      Scenario                      &  Method      & \thead{Attack \\ Acc\_top1↑} & \thead{Attack \\ Acc\_top5↑} & KNN\_Dist↓ & FID↓ & LPIPS\_Alex↓ & LPIPS\_VGG↓ \\
    \midrule
    \multicolumn{2}{c}{\multirow{4}{*}{Overlap}}    & \multirow{2}{*}{White-Box} & GMI    & 31.32\%    & 58.88\%    & 1105.4401 & \textbf{106.5101}  & 0.2421       & 0.4288      \\
\multicolumn{2}{c}{}                            &                            & PLG    & \textbf{57.88\%}    & \textbf{83.58\%}    & 1057.9677 & 153.9841   & 0.2524       & 0.4371      \\ \cmidrule(l){3-10}
\multicolumn{2}{c}{}                            & \multirow{2}{*}{Black-Box} & LB-MIA & 6.60\%     & 18.49\%    & 1334.4986 & 260.5981  & 0.3107       & 0.4332      \\
\multicolumn{2}{c}{}                            &                            & \textbf{Ours}   & 54.15\%    & 83.38\%    & \textbf{951.6782} & 112.1237  & \textbf{0.2057}       & \textbf{0.4151}      \\ \midrule
\multicolumn{2}{c}{\multirow{4}{*}{Nonoverlap}} & \multirow{2}{*}{White-Box} & GMI    & 30.80\%     & 58.66\%    & 1108.2195 & \textbf{102.7768}  & 0.2408       & 0.4264      \\
\multicolumn{2}{c}{}                            &                            & PLG    & \textbf{56.19\%}    & 82.94\%    & 1043.8163 & 131.2630  & 0.2333       & 0.4295      \\ \cmidrule(l){3-10}
\multicolumn{2}{c}{}                            & \multirow{2}{*}{Black-Box} & LB-MIA & 4.34\%     & 13.77\%    & 1383.2693 & 242.1953  & 0.3160       & 0.4384      \\
\multicolumn{2}{c}{}                            &                            & \textbf{Ours}   & 56.13\%    & \textbf{84.54\%}    & \textbf{959.3208} & 109.5714  & \textbf{0.2078}       & \textbf{0.4165}      \\
  \bottomrule
\end{tabular}}
\end{table*}

\begin{table*}
  \centering
  \caption{Quantitative evaluation and attack performance comparison on various methods under whether MNIST datasets overlap or not. ↑ and ↓ respectively symbolize that higher and lower scores give better attack performance.}
  \label{table6}
  \setlength{\tabcolsep}{4.7mm}{
\begin{tabular}{@{}cccccccc@{}}
\toprule
Dataset                     & Method        & \thead{Attack \\ Acc\_top1↑} & \thead{Attack \\ Acc\_top5↑ \\ (top2 in Nonoverlap)} & KNN Dist↓          & FID↓              & LPIPS\_Alex↓    & LPIPS\_VGG↓     \\ \midrule
\multirow{2}{*}{Overlap}    & LB-MIA        & 70\%              & 100\%             & 4274.1179          & 254.0187          & 0.4066          & 0.3582          \\
                            & \textbf{Ours} & \textbf{100\%}    & \textbf{100\%}    & \textbf{2634.9470} & \textbf{222.9036} & \textbf{0.2220} & \textbf{0.2954} \\ \midrule
\multirow{2}{*}{Nonoverlap} & LB-MIA        & 40\%              & 60\%              & 308.2348           & 278.2425          & 0.4047          & 0.4090          \\
                            & \textbf{Ours} & \textbf{76\%}     & \textbf{100\%}    & \textbf{146.9648}  & \textbf{238.6635} & \textbf{0.2816} & \textbf{0.3391} \\ \bottomrule
\end{tabular}%
}
\end{table*}
\subsubsection{Compare overlapping vs. non-overlapping auxiliary sets}
\begin{figure*}[t]
  \centering
  \includegraphics[width=0.92\linewidth]{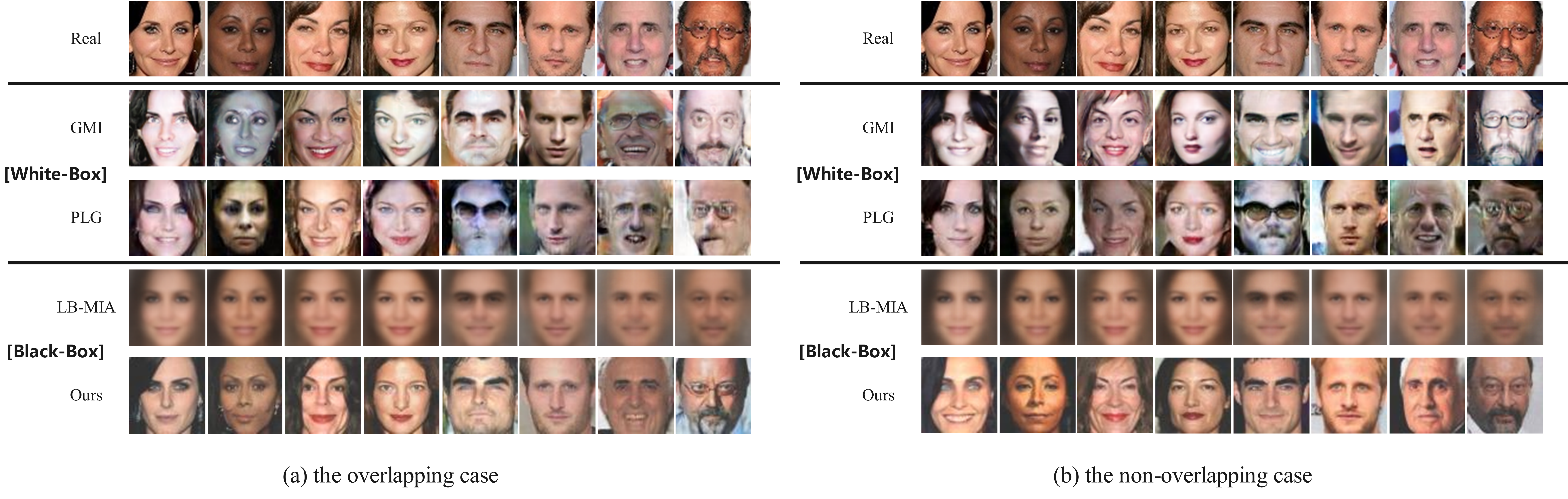}
  \caption{Qualitative evaluation and attack performance comparison on various methods under whether FaceScrub and CelebA datasets overlap or not.}
  \label{figure4}
\end{figure*}

We study for the first time whether the auxiliary set overlaps with the target training set on the impact of the attack. CelebA was chosen as the auxiliary set due to its greater distributional differences with FaceScrub compared to other face datasets, as established in previous work \cite{chen2021knowledge}\cite{yuan2023pseudo}\cite{kahla2022label}, and demonstrates that the selection of our dataset is the most challenging. While all previous studies assumed no individual overlap between the auxiliary and training sets by default, it is plausible for an adversary to obtain data for a small number of target individuals in real-world scenarios. Thus, it is meaningful to examine the effect of individual overlap on experimental results. As shown in Table \ref{table5}, our method achieves comparable results to those obtained with overlap even in the absence of individual overlap. However, there is a 2\% accuracy degradation between GMI and LB-MIA on these two auxiliary sets. The difference in accuracy between our method and the latest white-box attack PLG is approximately 1\%, which can be attributed to PLG’s reduction of generative image prediction and pseudo-labeling loss through the use of the target model during GAN training and optimization of the generator via gradient descent. But, in the non-overlapping case, the attack accuracy of our method is comparable to that of the PLG effect and even surpasses PLG under the top five accuracy evaluation. Furthermore, high accuracy is not the sole measure of a generative image’s effectiveness. In terms of feature distance and perceived similarity, our method yields the best results, producing recovered target image features that are more similar and realistic. From a quantitative evaluation perspective, our results ensure that recovered images closely resemble target individuals, with an average of 56.13\% of filtered results accurately identifying targets and 84.54\% of results indicating the presence of targets within the top five classes predicted by the model. Qualitative evaluation, as shown in Figure \ref{figure4}, further demonstrates that our methods generate more accurate and realistic results that closely resemble target individuals.

\begin{figure*}[t]
  \centering
  \includegraphics[width=0.88\linewidth]{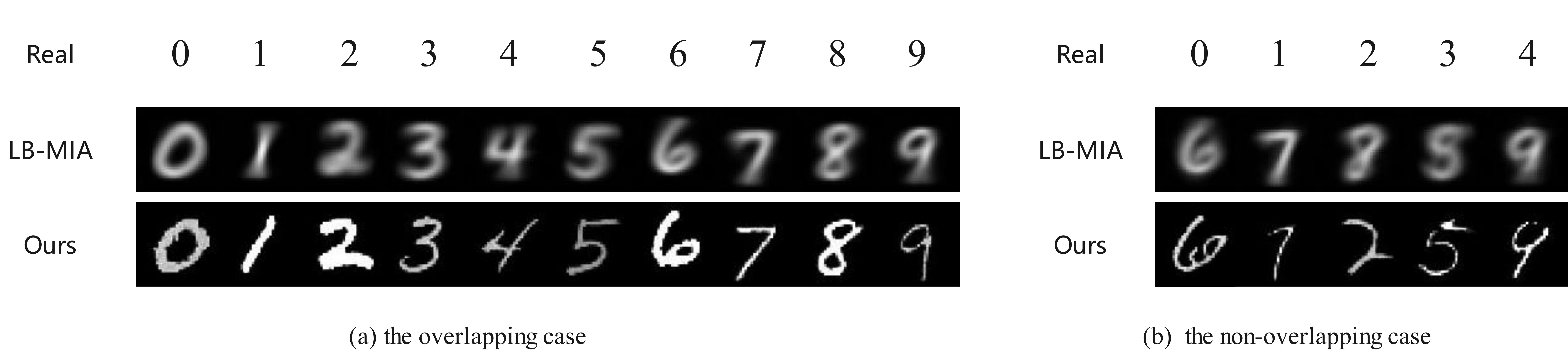}
  \caption{Qualitative evaluation and attack performance comparison on various methods under whether MNIST datasets overlap or not.}
  \label{figure5}
\end{figure*}

In addition to the recovery of face training set representative samples, we have also evaluated the MNIST dataset to demonstrate the generalization ability of this attack model. Due to the low complexity of this data, the inversion model capability of LB-MIA is sufficiently comparable to ours, so we only performed an equivalent comparison of the black-box generator for each metric. The quantitative evaluation is shown in Table \ref{table6}, where our method achieves an attack accuracy of 100\% when there is an overlapping in the dataset and is the best in all similarity metrics. The qualitative evaluation is shown in Fig. \ref{figure5}(a). When it is non-overlapping in the dataset, i.e., the target model is a 5-classification model, we evaluate the attack accuracy of top1 and top2, and again achieve the best in all metrics, with qualitative evaluation shown in Fig. \ref{figure5}(b). It should be noted that LB-MIA will only generate a unique image, and since the MNIST task is simpler than faces, generating multiple images would result in large differences in the FID metric. Therefore, we filtered only one of the most robust images in our method for evaluation in order to make a fair comparison.

Combined with the discussion in Section V and the results in Table \ref{table5}, it can be substantiated that the strategy of employing the generator in GAN as the attack model is subpar to our results in terms of comprehensive evaluation. This is even in the white-box condition due to the unstable training issue posed by the intricate loss function inherent to GAN. Furthermore, since PLG will utilize the loss between the predicted label of the generative image and the target label to optimize the generator, it results in feature similarity and sensory similarity that is not superior to our method, even though the ultimate result emphasizes ensuring attack accuracy. The training of the conditional diffusion model, which is optimized with the loss between the prediction noise and the actual noise, is more stable compared to GAN. Owing to the guidance of predicted labels, it can guarantee the accuracy and structural authenticity of the image. Moreover, a limitation of the attack model in LB-MIA is that it only generates a single image for each label, which reduces the fault tolerance of the attack. However, our trained attack model can generate multiple possible images in label-only scenarios and filter them with the assistance of the target model.

Based on the above analysis, it can be observed that both KNN Dist and FID have limitations. For instance, a review of Table \ref{table6} reveals that KNN Dist exhibits significant variability in results under different evaluation models, which poses challenges for comparative experimental evaluations within the field. Furthermore, since the model used for evaluating the KNN Dist was trained on the same training set as the target model, and the FID was only calculated based on the pre-trained Inception v3, the FID is not more reliable than the KNN Dist in different scenarios. Consequently, FID is gradually being phased out for evaluations within the MIA field. On the other hand, the LPIPS evaluation metric, which is based on a pre-trained model, quantifies the perceptual similarity between images. As demonstrated in Table \ref{table5}, \ref{table6}, and Figure \ref{figure4}, LPIPS exhibits greater generalizability than KNN Dist, i.e., it maintains a consistent value domain, thereby facilitating comparative experimental evaluations within the field. In comparison to FID, LPIPS proves to be more reliable. For example, referring to Table  \ref{table5}, it can be observed that the results generated by our method outperform GMI in terms of attack accuracy, feature distance, and qualitative assessment. However, GMI outperforms all methods under the FID. Referring to Table  \ref{table6}, it can be seen that our results exhibit a larger gap in attack accuracy and feature distance compared to LB-MIA, but FID does not accurately reflect this. In such instances, the quantitative value of LPIPS provides a more accurate representation of the effect gap between different methods.

\subsubsection{Evaluate attack performance for same target label}
\begin{figure*}[t]
  \centering
  \includegraphics[width=0.88\linewidth]{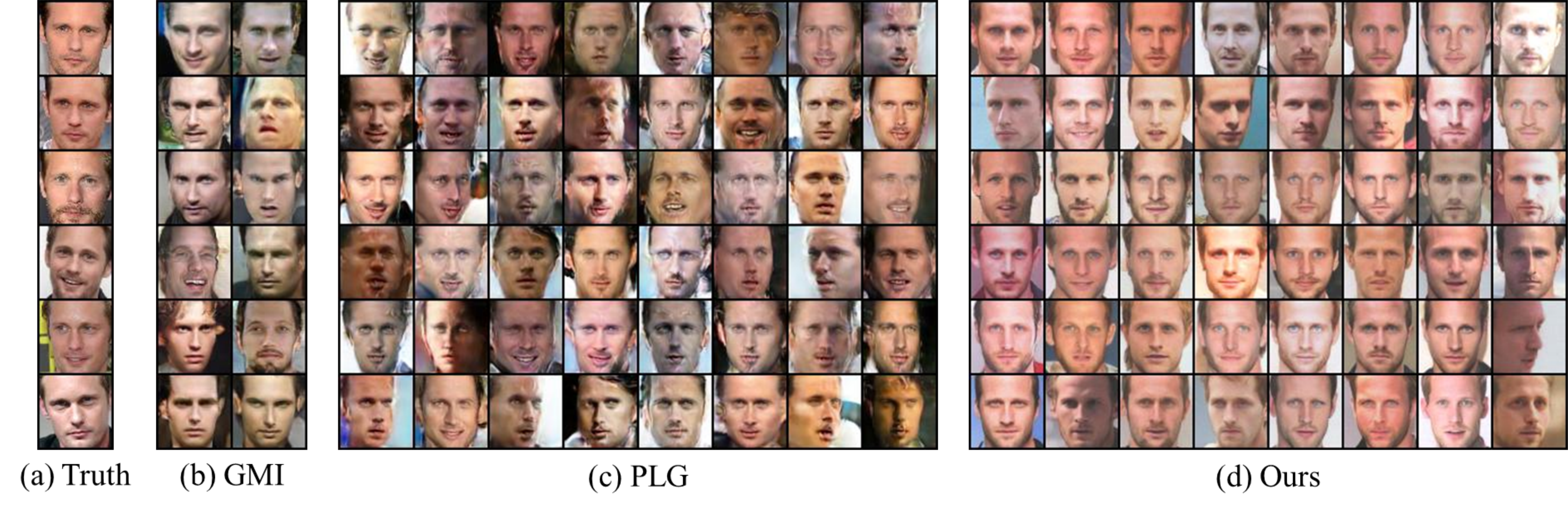}
  \caption{Qualitative evaluation and attack performance comparison on various methods under the same target label \textit{when there is no individual overlap}. (a) represent the real target images, (b) shows the GMI attack result, (c) displays the PLG attack result, and (d) presents the attack result of our method. Unlike (b), (c, d) are recovered directly from input noise and a label without optimization and filtering.}
  \label{figure6}
\end{figure*}

According to the results presented in Table \ref{table5}, it can be observed that the generator trained using a white-box attack compromises generation quality in favor of improved accuracy. This attack relies on the optimization algorithm employed during the recovery phase, with the essence of the optimization study being equivalent to replacing the recovery phase optimization algorithm in GMI for both black-box and white-box attacks. However, this may prove challenging in real-world scenarios. Our focus is on assessing the security risk posed by a powerful attack generator, as opposed to previous work that concentrated on studying optimization algorithms. As shown in Figure \ref{figure6}, our method is capable of generating results without the need for optimization and filtering, yielding recovery results that are more realistic and closely resemble target individuals compared to those generated by other methods. In addition, it can also be observed from the recovery results for the same label that the results recovered by our method are more compact. Moreover, such a comparison provides a fairer representation of the capabilities of diffusion models relative to GANs.

\begin{figure*}[h]
  \centering
  \includegraphics[width=0.8\linewidth]{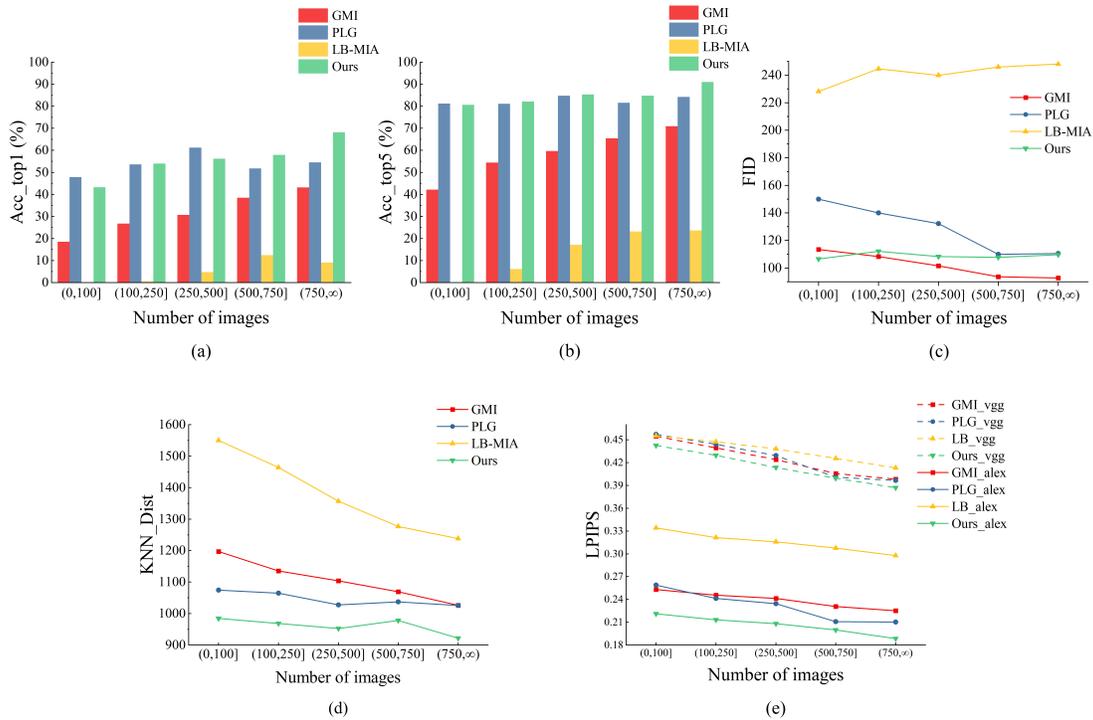}
  \caption{Quantitative evaluation and attack performance comparison on various numbers of auxiliary images \textit{when there is no individual overlap}. (a, b) represent the effect of quantity on accuracy, (c, d) represent the effect of quantity on similarity, and (e) represent the effect of quantity on similarity and realism.}
  \label{figure7}
\end{figure*}

\vspace{-0.1cm}

\subsubsection{Examine the impact of auxiliary data quantity on attack}
Our study analyzes both the generator and the auxiliary set and compares the results obtained with different quantities of data assigned to the target individual within the auxiliary set. As shown in Figure \ref{figure7}, our findings indicate an overall upward trend in attack accuracy as the number of auxiliary sets increases. In contrast to PLG, our approach demonstrates a consistent upward trend and surpasses the other three methods in attack accuracy when provided with a substantial quantity of auxiliary data. Specifically, with over 750 auxiliary data, top-1 accuracy reaches 68.18\%, significantly surpassing the attack accuracy of PLG. Top-5 accuracy attains 84.82\% and 91.06\% for quantities ranging from 500 to 750 and above, respectively, both exceeding that of PLG. 

Figure \ref{figure7} (c-e) shows that the feature distance of recovery results for each method gradually decreases as the quantity increases, with our method achieving the best evaluation in terms of similarity. Furthermore, since both KNN Dist and Attack Acc are computed by the same evaluation model. Theoretically, only the fully connected layer of the evaluation model affects the results of both. By examining (a, b, d), it can be observed that KNN Dist exhibits a trend similar to the attack accuracy. For instance, both PLG and our results show a decreasing trend when the number is in the interval (500,750] in (b), which can also be visualized in (d). Compared to the trends presented by FID and KNN Dist, the evaluation of LPIPS is more objective. We consider a quantitative trend presentation for the qualitative evaluation and demonstrate the trend that the quality of the generation progressively improves as the number increases. Comparative analysis reveals that, with an equivalent amount of data, our approach attains a superior recovery level.

\begin{table*}
\centering
  \caption{Quantitative evaluation and effect of gamma factor $\gamma$ and guidance strength $\omega$ on attack performance.}
  \label{abalation}
  \setlength{\tabcolsep}{5.2mm}{
  \begin{tabular}{cccccccc}
\toprule
Factor                 & Value    & \thead{Attack \\ Acc\_top1↑}       & \thead{Attack \\ Acc\_top5↑}       & KNN\_Dist↓    & FID↓    & LPIPS\_Alex↓    & LPIPS\_VGG↓     \\ \midrule
\multirow{7}{*}{$\gamma$} & 1   & 38.32\%          & 69.17\%          & 1050.698    & 124.9308      & 0.2470          & 0.4325          \\
                        & 2        & 54.29\%          & 83.6\% & 965.8551 & \textbf{109.1064} & 0.2081          & \textbf{0.4161}          \\
                        & 2.1      & 55.33\%          & 83.96\%          & 963.0374     & 109.1552     & 0.2077          & \textbf{0.4161}          \\
                        & 2.2      & 55.9\% & 84.26\%          & 961.1259   & 109.1554       & \textbf{0.2076} & 0.4164          \\
                        & 2.3      & 56.13\%          & 84.54\%          & 959.3208     & 109.5714     & 0.2078          & 0.4165          \\
                        & 2.4      & 56.24\%          & \textbf{84.76\%}          & 958.8945    & 109.5858      & 0.2081          & 0.4168          \\
                        & 2.5 & \textbf{56.45\%}          & 84.72\%          & \textbf{958.3543}   & 109.9230       & 0.2086          & 0.4172          \\ \midrule
\multirow{5}{*}{$\omega$} & 1      & 33.89\%          & 63.89\%          & 1009.7654     & \textbf{94.3645}
     & \textbf{0.2024} & \textbf{0.4078} \\
                        & 2   & 41.53\%          & 74.72\%          & 984.6321    & 98.3713
      & \textbf{0.2024} & 0.4082          \\
                        & 4   & 55.00\%          & \textbf{84.72\%}          & \textbf{945.4405}   & 108.7672
       & 0.2060          & 0.4185          \\
                        & 6   & 56.25\% & 83.75\% & 955.6436 & 119.0539
 & 0.2134          & 0.4257          \\
                        & 8   & \textbf{57.92\%}          & 84.44\%          & 966.1134     & 136.4142
     & 0.2233          & 0.4412          \\ \midrule

 \multirow{4}{*}{$p$}& 0& 44.4\%& 76.67\%& \textbf{768.6025}& 119.4697& 0.2068&\textbf{0.4171}\\
 & 0.1& \textbf{55.00\%}& \textbf{84.72\%}          & \textbf{945.4405}& \textbf{108.7672}& \textbf{0.2060}&0.4185          \\
 & 0.2& 42.36\%& 74.17\%& 955.4660& 115.9034& 0.2153&0.4267\\ 
 & 0.3& 42.78\%& 71.25\%& 962.8414& 115.9907& 0.2145&0.4273\\ \bottomrule
\end{tabular}}
\end{table*}

\subsubsection{Examine the impact of different target models on attack}
The predictive ability of the target model directly determines the effectiveness of this attack. For example, if a face recognition model can better determine the closest training set individual with that feature from the auxiliary data, then the features guided by the same label in training will be closer and the attack will be more effective. In this section, we only discuss the impact of different target model capabilities on this method. Because the effect of different model architectures on white-box attacks is not the purpose of this paper's discussion. In addition, there is no need to repeat the LB-MIA experiment because is too ineffective. As shown in Table \ref{table4} and \ref{dtm}, as the target model prediction capability increases, the evaluation of each metric is better, and the accuracy of the top-5 attack on IR152 reaches 93.76\%.

When the target model is CNN and VGG16, metrics such as attack accuracy, KNN Dist, and FID can reflect some degree of effect enhancement. However, the evaluation by LPIPS indicates that the result may not show significant improvement from the perspective of human-eye sensory similarity. When the target model is IR152, which possesses the strongest capability, the attack only has an effect on sensory similarity. The enhancement of the attack effect can also be observed from the FID result.

\begin{table}[t]
\centering
  \caption{Quantitative evaluation and effect of different target models on attack performance. ↑ and ↓ respectively symbolize that higher and lower scores give better attack performance.}
  \label{dtm}
  \setlength{\tabcolsep}{1.34mm}{
\begin{tabular}{@{}ccccccc@{}}
\toprule
\multicolumn{1}{l}{\thead{Target \\ model}} & \thead{Attack \\ Acc\_top1↑} & \thead{Attack \\ Acc\_top5↑} & \thead{KNN \\ Dist↓ }        & FID↓              & \thead{LPIPS 
\\ \_Alex↓}    & \thead{LPIPS \\ \_VGG↓}     \\ \midrule
CNN                              & 56.13             & 84.54             & 959.3208          & 109.5714          & 0.2078          & 0.4165          \\
VGG16                            & 66.6              & 90.36             & 893.884           & 108.2697          & 0.2075          & 0.4128          \\
\textbf{IR152}                   & \textbf{74.8}     & \textbf{93.76}    & \textbf{855.1517} & \textbf{101.0693} & \textbf{0.2015} & \textbf{0.4042} \\ \bottomrule
\end{tabular}%
}
\end{table}

\subsubsection{Ablation study}

\begin{figure}[t]
  \centering
  \includegraphics[width=0.95\linewidth]{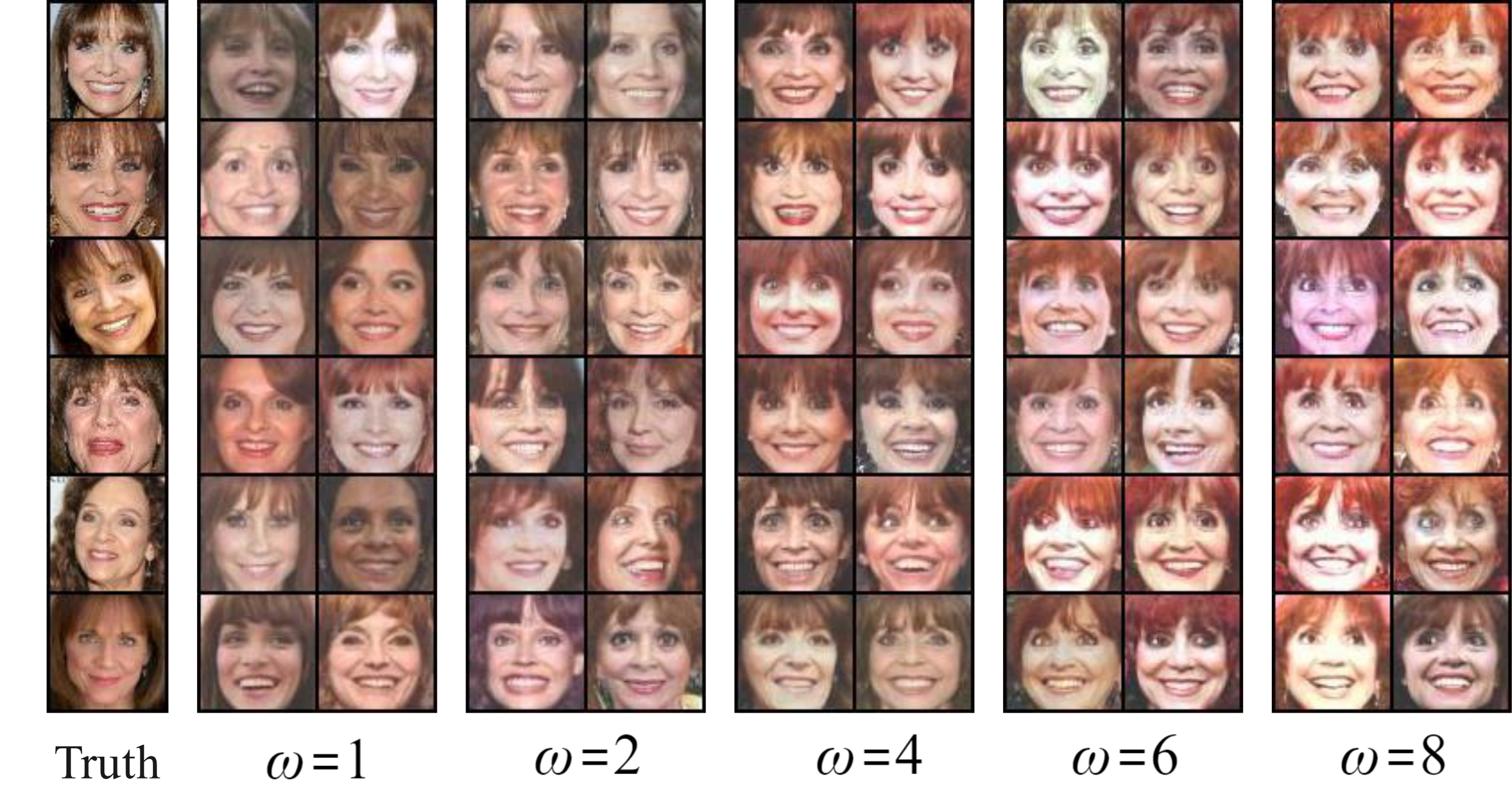}
  \caption{Qualitative evaluation and effect of guidance strength $\omega$ on attack performance.}
  \label{figure8}
\end{figure}

We conducted further analysis on the effect of the gamma factor \begin{math}
    \gamma
\end{math} and guidance strength \begin{math}
    \omega
\end{math} on our experimental results. As depicted in Table \ref{abalation} our findings confirm that an increase in \begin{math}
    \gamma
\end{math} positively impacts attack accuracy and KNN distance, where \begin{math}
    \gamma=1 
\end{math} indicates the absence of image correction. However, for perceptual similarity, the evaluated results exhibit a decreasing and then increasing trend and the $\gamma$ equal of 2.2 is the most consistent with human perceptual judgment, necessitating a trade-off in the value of gamma. Second, as the value of \begin{math}
    \omega
\end{math} increases, there is an initial improvement followed by a decline in both the evaluation of top-5 attack accuracy and feature distance, while perceptual similarity progressively becomes worse. Since when \begin{math}
    \omega=0
\end{math} indicates unconditional guidance generation, it is not explicitly evaluated in the results. As shown in Figure \ref{figure8}, excessively high values of \begin{math}
    \omega
\end{math} result in a degradation of image quality but in favor of target features. However, as can be observed from the figure, a lower KNN Dist does not necessarily imply a more effective attack. Similarly, under the influence of $\gamma$ and $\omega$, the evaluation of FID, when compared to that shown by LPIPS, reveals that the results of LPIPS are more closely aligned with the qualitative judgment of the human eye. This aids in the selection of hyper-parameters and the evaluation of the results. So, in order to effectively balance accuracy, realism, and similarity, it is crucial to select an appropriate level of guidance strength with reference to the LPIPS. For instance, when $\omega$ equals 4, the overall evaluation results exhibit greater balance.

Furthermore, we delve into the impact of the probability $p$ of training without label guidance on the results. This hyper-parameter is introduced as it allows for the learning of the data’s features and structure with a certain probability, thereby ensuring that the attack model does not over-fit the label information. When $p$ is 0, indicating that the attack model is entirely guided by labels, it is observed that the feature distance of KNN Dist is evaluated to be 768.6025, which is the lowest. However, this is not the optimal choice in terms of attack accuracy evaluation. When $p$ exceeds 0.2, a diminishing effect can be observed through the evaluation of all indicators. Therefore, for the best effect, $p$ should take a value between 0 and 0.2.

\vspace{-0.1cm}

\section{Limitations and Future Work}
Despite achieving favorable experimental results in our proposed setting, our method has limitations when considered alongside current related works.

\begin{enumerate}
    \item Training the conditional diffusion model takes longer compared to other methods due to the multiple rounds of iterations required to learn the noise distribution. Research on accelerating diffusion model training is ongoing and this limitation may be overcome in the future.

    \item We focus on developing a robust and practical attack model in label-only scenarios. However, secondary optimizations, based on the results produced by this model, may offer more opportunities for enhancement than the outcomes of existing generators. For instance, the optimization of the initial input noise or the images in the noise reduction path. We believe this is a direction deserving of further investigation in future research.
\end{enumerate}

Furthermore, designing defense strategies against label-only MIAs will be a challenging direction. We believe that future “model-centric" defense strategies\cite{wang2021improving}\cite{peng2022bilateral} should effectively leverage powerful deep learning models available today. “Data-centric” defense strategies\cite{struppek2023careful}\cite{chendata} need to trade off the prediction accuracy of the target model for training and non-training sets.

In conclusion, it is imperative for future AI research to prioritize the protection of privacy knowledge acquired by models while enhancing their utility. This is particularly relevant in light of recent developments in model inversion attacks and the rapid advancement of AI technology.

\section{Conclusion}
We develop a novel label-only model inversion attack method utilizing a conditional diffusion model, capable of recovering representative data for a specific target label in the training set, given that the target model predicts only the label for the input. The attack model is trained on an auxiliary public dataset and uses the predicted label of corresponding auxiliary data as a condition to guide the training of the diffusion model. This allows the adversary to input standard normally distributed noise and the target label into the conditional diffusion model during the recovery phase, generating data with a pre-defined guidance strength representing that label in the training set without optimization. Experimental results demonstrate that our method generates more accurate, realistic, and similar data compared to generators in related work. Future work will focus on exploring more efficient optimization algorithms based on such a high-quality generator and investigating defense methods that balance model privacy and usability.

\section*{ACKNOWLEDGMENTS}
 This work was supported by the “Pioneer” and “Leading Goose” R\&D Program of Zhejiang (Grant No.2023C03203, 2023C03180, 2022C03174), and the Key Science and Technology Research Project of Chongqing Municipal Education Commission, KJZD-K202300601.

\bibliographystyle{IEEEtran}
\bibliography{sample-sigconf}

\end{document}